\documentclass[sigconf]{acmart}

\usepackage{algorithmic}
\usepackage{svg}
\usepackage{listings}
\usepackage{float}
\usepackage{needspace}
\usepackage{xcolor}

\definecolor{codebg}{RGB}{245,245,250}    
\definecolor{codeframe}{RGB}{200,200,220} 
\definecolor{stringcolor}{RGB}{120,160,120} 
\definecolor{keywordcolor}{RGB}{80,120,180} 
\definecolor{commentcolor}{RGB}{150,150,150} 

\AtBeginDocument{%
  }

\setcopyright{acmlicensed}
\copyrightyear{2025}
\acmYear{2025}
\acmDOI{XXXXXXX.XXXXXXX}

\acmConference[DAI '25]{The Seventh International Conference on Distributed Artificial Intelligence}{Nov 21 - 24, 2025}{London, United Kingdom}
\acmBooktitle{Proceedings of The Seventh International Conference on Distributed Artificial Intelligence (DAI'25), Nov 21 - Nov 24, 2025, London, United Kingdom}
\acmISBN{978-1-4503-XXXX-X/25/11}

\begin{document}

\title{Learning to Play Blackjack: A Curriculum Learning Perspective}

\author{Amirreza Alasti}
\affiliation{%
  \institution{Leibniz University Hannover}
  \city{Hannover}
  \country{Germany}}
\email{amirreza.alasti@stud.uni-hannover.de}

\author{Efe Erdal}
\affiliation{%
  \institution{Leibniz University Hannover}
  \city{Hannover}
  \country{Germany}}
\email{efe.erdal@stud.uni-hannover.de}

\author{Yücel Celik}
\affiliation{%
  \institution{Leibniz University Hannover}
  \city{Hannover}
  \country{Germany}}
\email{yucel.celik@stud.uni-hannover.de}

\author{Theresa Eimer}
\affiliation{%
  \institution{Leibniz AI Academy}
  \city{Hannover}
  \country{Germany}}
\email{t.eimer@ai.uni-hannover.de}

\renewcommand{\shortauthors}{Alasti, et al.}

\begin{abstract}
Reinforcement Learning (RL) agents often struggle with efficiency and performance in complex environments. We propose a novel framework that uses a Large Language Model (LLM) to dynamically generate a curriculum over available actions, enabling the agent to incorporate each action individually. We apply this framework to the game of Blackjack, where the LLM creates a multi-stage training path that progressively introduces complex actions to a Tabular Q-Learning and a Deep Q-Network (DQN) agent. Our evaluation in a realistic 8-deck simulation over 10 independent runs demonstrates significant performance gains over standard training methods. The curriculum-based approach increases the DQN agent's average win rate from 43.97\% to 47.41\%, reduces the average bust rate from 32.9\% to 28.0\%, and accelerates the overall workflow by over 74\%, with the agent's full training completing faster than the baseline's evaluation phase alone. These results validate that LLM-guided curricula can build more effective, robust, and efficient RL agents.
\end{abstract}



\maketitle

\section{Introduction}
Reinforcement Learning (RL) is a rapidly evolving field with applications ranging from robotics~\cite{lee-sciro20a,reinforcement_in_robotics} to game playing~\cite{mnih_human-level_2015,vasco-rlc24,Li_2025} and, more recently, large language models (LLMs)~\cite{ouyang-neurips22a,RL_LLM}. As RL methods are increasingly applied in complex and uncertain environments~\cite{bakhtin-iclr23,matthews-icml24,imrie2025assuringsafetyreinforcementlearning}, it becomes crucial to understand their limitations and to identify strategies for improving learning efficiency and generalization. The efficacy of RL is often hindered by two fundamental challenges: vast state-action spaces that make thorough exploration intractable (a phenomenon known as the "curse of dimensionality"~\cite{lu2024overcomingcursedimensionalityreinforcement,chen2021infinitehorizonofflinereinforcementlearning}), and sparse rewards that provide insufficient feedback to guide the learning process~\cite{pmlr-v97-agarwal19e}. These issues can lead to unstable policies and inefficient learning, motivating the need for structured training strategies.

To address these challenges, Curriculum Learning has been proposed as a strategy to guide the agent through a structured sequence of tasks, starting from simpler ones and gradually increasing in difficulty~\cite{curriculum_learning,CR_RL_SUR}. By doing so, the agent can incrementally build the necessary representations and policies required to handle the full environment more effectively.

Building such curricula, however, is often non-trivial.
Many curriculum learning approaches rely on hand-crafting task sequences or at least manually defining simple start and difficult goal tasks~\cite{klink-icml22}.
Where these are unknown, unsupervised methods can use agent performance as a rough guide~\cite{portelas-corl19,eimer-icml21,parkerholder-icml22}. 
With the increasing availability of LLMs, however, they have been proposed as an accessible replacement for outside expertise\cite{wang2023voyager,ma-iclr24,liang2024environment}.
Our approach progressively increases task complexity based on learning progress via an LLM-generated curriculum across available actions.
This allows the agent to incorporate one action at a time into its policy according to the LLM's understanding of the task. 

Blackjack serves as an excellent domain for this study, not only due to its partially observable state but also because it has a well-defined performance benchmark. For human players, adhering to an optimal "basic strategy" is critical. Under typical casino rules, a player using perfect basic strategy can reduce the house edge to as low as 0.5\%~\cite{griffin1999theory, thorp1966beat}. This translates to the player winning approximately 42.4\% of hands, excluding ties~\cite{shackleford_blackjack}. In contrast, a casual player who deviates from this strategy often faces a much larger house edge of 2\% or more. Therefore, a key measure of our RL agent's success is not just its ability to learn, but its capacity to converge on a policy that meets or exceeds this expert human-level performance, effectively neutralizing the inherent house advantage.

With this motivation in mind, our research is guided by the following questions: Can Curriculum Learning influence the performance of RL agents in a task like Blackjack? And how does increasing environmental complexity, such as simulating more realistic scenarios through larger deck sizes, affect their learning behavior?
To explore these questions, we investigate the performance and limitations of two widely used RL algorithms: Tabular Q-learning \cite{watkins_q-learning_1992} and Deep Q-Networks (DQN) \cite{dqn} within the Blackjack environment. We conduct experiments both with and without Curriculum Learning based on Google Gemini 2.0 Flash to analyze how LLM-guided curricula impact convergence, generalization, and policy quality as task difficulty increases. Our code, including full results, can be found at \url{https://github.com/LUH-AI-devnerds/llm-guided-curriculum-rl}. 

\paragraph{\textbf{Contributions.}} This paper makes the following contributions:
\begin{itemize}
    \item \textbf{LLM-guided action curriculum:} an LLM-guided staged introduction of actions 
    with \emph{adaptive success thresholds} driven by compact performance summaries.
    \item \textbf{Empirical study in the Blackjack environment:} a thorough evaluation of our action curriculum for DQN and Tabular agents on Blackjack, showing an up to $3.4\%$ increase in win rate, $4.9\%$ decrease in bust rate, while accelerating training by $74\%$
    \item \textbf{Strategy analysis:} an in-depth qualitative analysis of agent strategies and how they relate to curriculum stages.
\end{itemize}

\section{Related Work}
Our research is situated at the intersection of three areas we introduce in this Section: Reinforcement Learning applications in Blackjack, Curriculum Learning methodologies, and the emerging use of Large Language Models to guide RL agents~\cite{10766898, pmlr-v202-du23f, ryu2025curricullmautomatictaskcurricula}.

\paragraph{\textbf{Reinforcement Learning in Blackjack.}}
Blackjack has long been a standard testbed for RL algorithms due to its well-defined rules and non-trivial strategic depth~\cite{thorp1966beat}. Similar research has successfully applied a range of techniques, including Monte Carlo methods~\cite{james1980monte}, Temporal-Difference (TD) learning~\cite{tesauro1995temporal}, and tabular Q-learning~\cite{watkins1992q}, to find effective playing strategies~\cite{srinivasaiah2024reinforcement, deGranville}. These studies typically frame the game as a finite Markov Decision Process (MDP)~\cite{puterman1990markov} and demonstrate that model-free agents can converge to policies that approximate optimal basic strategy. Other work focuses on comprehensive simulations to benchmark various approaches, from hand-crafted basic strategy charts and card-counting systems to different ML-based models, comparing them on metrics like win rate and return on investment~\cite{ali2025evaluating}.

However, a common thread in these foundational works is the immediate exposure of the agent to the \emph{entire} action space from the beginning of training. This approach, while direct, presents a significant exploration challenge. The action space includes not only frequent decisions (Hit, Stand) but also rare, context-dependent, and powerful actions (Split, Double Down)~\cite{sutton1998reinforcement}. Exposing all actions at once increases the branching factor and can make it difficult for the agent to assign credit correctly, potentially slowing convergence or leading to suboptimal policies that underutilize crucial tactical moves. Our work addresses this gap through a structured introduction of actions.

\paragraph{\textbf{Curriculum Learning for Reinforcement Learning.}}
Curriculum Learning (CL) proposes reframing a complex learning problem into a sequence of progressively harder sub-tasks~\cite{curriculum_learning}. The core benefit of a curriculum is that it guides the agent's exploration, allowing it to build foundational knowledge on simpler tasks before tackling the full complexity of the environment. This often leads to improved sample efficiency, better generalization, and more stable convergence, especially in environments with sparse rewards or high-dimensional state-action spaces. In RL, this principle has been applied in various forms~\citep{CR_RL_SUR}, such as starting with a simplified state space~\citep{wohlke2020performance}, gradually increasing the difficulty of goals~\citep{portelas-corl19,klink-icml22}, shaping rewards over time~\citep{8539438}, or introducing adversaries of increasing skill in self-play scenarios~\cite{narvekar2020curriculum}. For Blackjack, an action-based curriculum is a natural fit, mimicking how human players learn: master hitting and standing before tackling the nuances of doubling down and splitting pairs.

\paragraph{\textbf{LLMs as Curriculum Generators.}}
More recently, Large Language Models (LLMs) have emerged as powerful tools for automating the design of learning curricula. This paradigm shifts the burden of manual curriculum engineering to the reasoning capabilities of an LLM. Groundbreaking work like Voyager~\cite{wang2023voyager} uses GPT-4 as an unbounded task generator for open-ended exploration in Minecraft, creating a self-driven curriculum that continuously expands an agent's skills. Similarly, Eureka~\citep{ma-iclr24} and Eurekaverse~\cite{liang2024environment} leverage LLMs to co-evolve agents and their training environments to master complex robotic motor skills.

Our work explores a different paradigm from these open-ended systems. We use the LLM as a \emph{domain-specific pedagogical expert} for a well-defined task. Instead of inventing new goals, the LLM is prompted to structure the learning process based on a high-level understanding of action complexity in Blackjack. By providing it with compact performance summaries from our agent, we prompt the LLM to act as an automated 'coach,' deciding when the agent is ready for the next level of strategic complexity. To our knowledge, this is the first systematic application of an LLM-guided \emph{action curriculum} to the domain of Blackjack. It directly addresses the exploration challenges inherent in classic RL approaches to the game by bridging them with the reasoning capabilities of modern LLMs. The following sections will demonstrate that this structured methodology not only produces a final policy of superior quality and efficiency but also yields key insights into the learning process itself, such as identifying an optimal depth for the curriculum.

\section{Methodology}
Our LLM-guided curriculum learning framework systematically structures the agent's learning process to improve training efficiency and final policy quality. The core principle is to decompose the complex task of learning Blackjack by creating a curriculum over the available actions. Instead of confronting the agent with the entire action space from the start, the LLM generates a sequence of stages with progressively increasing complexity.

\begin{figure*}[t]
    \centering
    \includegraphics[width=\textwidth]{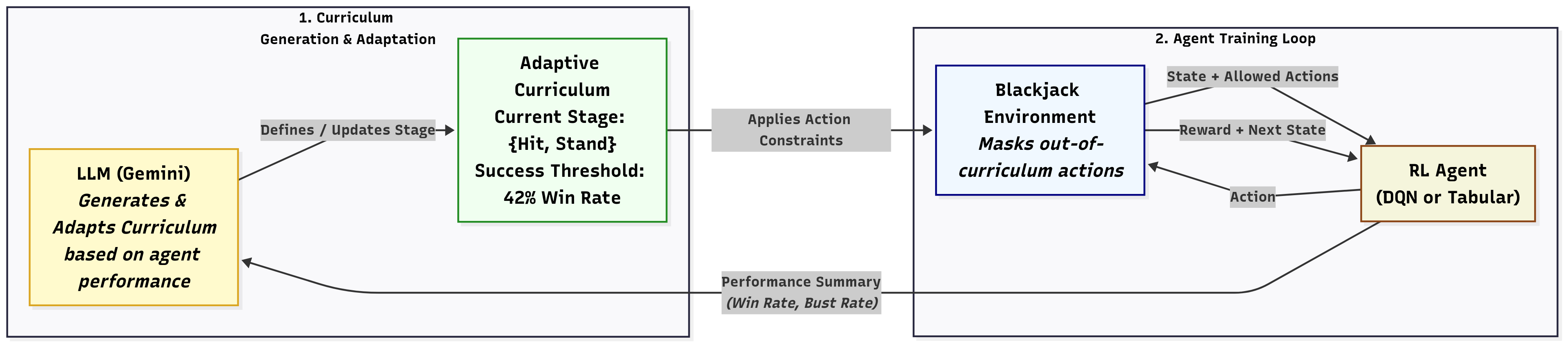} 
    \caption{The adaptive, LLM-guided curriculum learning framework. 
    \textbf{(1) Generation \& Adaptation:} An LLM generates an initial curriculum stage. After a training phase, the agent's performance summary is fed back to the LLM, which adapts the curriculum by deciding whether to advance the agent to the next stage. 
    \textbf{(2) Training Loop:} The agent's training is constrained by the current curriculum stage, which masks unavailable actions in the environment to focus exploration.}
    \Description{A flowchart showing an adaptive loop. In phase 1, an LLM generates a curriculum. In phase 2, an RL agent trains in an environment constrained by that curriculum. A feedback arrow shows that the agent's performance summary from phase 2 is sent back to the LLM in phase 1, which then updates the curriculum.}
    \label{fig:system_architecture}
\end{figure*}

In each stage, the agent is restricted to a specific subset of actions, allowing it to master simpler strategies (e.g., Hit/Stand) before moving on to more nuanced ones (e.g., Double Down, Split). The LLM is responsible for defining these action subsets and setting an adaptive performance threshold, a target win rate that the agent must achieve to advance to the next stage. 
Figure \ref{fig:system_architecture} illustrates the workflow from LLM guidance through curriculum stages to our multi-agent training setup.

\subsection{LLM-Guided Curriculum Design}
The curriculum generation utilizes the Google Gemini API, specifically the \textbf{Gemini 2.0 Flash} model, to dynamically create progressive learning stages. The LLM receives structured prompts containing six available actions which are: "Stand", "Hit", "Double Down", "Split", "Early Surrender" and "Insurance" with manually defined complexity levels 1-3, where Stand and Hit are considered as complexity level 1, Double Down and Split as level 2 and Early Surrender and Insurance as level 3. 
This complexity ranking is based on the actions' roles in the game. Stand and Hit are the basic game options, Double Down and Split are additional strategies, and Surrender and Insurance have further-reaching consequences like ending the game. The system then generates curriculum stages with adaptive success thresholds based on the win rate and progressive action introduction based on complexity.

\subsection{LLM Prompting and Curriculum Workflow}
\label{sec:llm_workflow}
Our adaptive curriculum is driven by a structured workflow that cycles between curriculum generation, agent training, and performance-based adaptation. This process is orchestrated through carefully designed prompts to the LLM.

\paragraph{\textbf{Initial Curriculum Generation.}}
The process begins by prompting the Gemini 2.0 Flash model to generate an initial multi-stage curriculum. We use conservative decoding settings (temperature 0.2, top-$p$ 0.9) to ensure consistent, structured output. The initial prompt provides the LLM with the complete context of the task, including: (i) the environment configuration (e.g., 8-deck), (ii) the full set of available actions grouped by complexity, and (iii) an explicit JSON format that the output must follow. This format defines the structure for each stage, specifying its name, the subset of available actions, a descriptive goal, and a target success threshold. An example of this prompt is shown in Listing~\ref{lst:llm_prompt}.

\begin{minipage}{\linewidth}
\begin{lstlisting}[language=bash, caption={Example LLM prompt for curriculum generation and adaptation. The `Last stage summary` field is used in the feedback loop.}, label={lst:llm_prompt}, breaklines=true, basicstyle=\small\ttfamily, frame=single]
SYSTEM: You are designing a Blackjack learning curriculum.
USER:
Environment: deck=8-deck, penetration=0.9
Actions: 0=Stand,1=Hit,2=Double,3=Split,4=Surrender,5=Insurance
Complexity: {1:[0,1], 2:[2,3], 3:[4,5]}
Last stage summary: {id:3, win_rate:0.455, bust_rate:0.31,
  errors:["over-hitting hard 15 vs 10"]}

Return ONLY JSON with fields:
{"stage_id": int, "name": str, "available_actions": [ints],
 "description": str, "difficulty": int [1..5],
 "success_threshold": float in [0.35, 0.50]}
\end{lstlisting}
\end{minipage}

\paragraph{\textbf{Adaptive Training Loop.}}
Once the curriculum is established, the agent begins training in Stage 1. The core of our framework is an adaptive loop that uses a progressive training budget for each stage. The number of episodes allocated per stage increases with its difficulty to provide more training time for complex actions. Performance is evaluated, and the feedback loop with the LLM is triggered when one of two conditions is met: either the agent achieves the stage's predefined success threshold (e.g., a target win rate) or it exhausts its allocated episode budget for that stage (capped at a maximum of 100,000 episodes). At this point, a compact performance summary, including metrics like win rate and common errors (as seen in Listing~\ref{lst:llm_prompt}), is sent back to the LLM. The LLM is then prompted to act as a coach, deciding whether the agent is ready to advance or if it needs to continue training. This feedback mechanism allows the LLM to dynamically manage the agent's progression.

\paragraph{\textbf{Robustness and Enforcement.}}
To ensure reliable operation, the system incorporates several robustness measures. The LLM's JSON responses are strictly validated against the required format. In case of malformed output or API failures, a retry mechanism is triggered; if the issue persists, the system can fall back to a pre-generated curriculum file to ensure experiments can run offline. Throughout our experiments, this fallback mechanism was never triggered. The combination of structured prompting and the retry logic was sufficient to handle any intermittent API issues, demonstrating the high reliability of the LLM for this structured generation task. Within the training loop, the curriculum is enforced by the environment, which masks any actions that are not part of the current stage. The environment also enforces all game rules (e.g., disallowing a Double Down after the first two cards), which prevents the agent from learning invalid policies and simplifies credit assignment.

\subsection{Blackjack Environment}
We implement Blackjack with full casino rules with multiple deck configurations (infinite, 1-deck, 4-deck, 8-deck) and configurable penetration. 
The dealer's behavior is encoded in the environment, and the game seats are occupied by separate RL agents. We play with two agents in our experiments.

\paragraph{\textbf{Observations}} The observation space is designed to provide the agent with all necessary information for strategic play. It includes the player's current total, the dealer's visible up-card, and a set of binary flags. These flags specify: \textbf{(i)} if the player's hand is "soft"—meaning it contains an Ace that can be counted as 11 without busting, a crucial distinction for optimal strategy—and \textbf{(ii)} which tactical actions (split, double, surrender, insurance) are currently legal to play. For experiments with finite decks, we also provide card-counting features (the running count and true count) to allow the agent to learn policies that adapt to the composition of the remaining shoe.

\paragraph{\textbf{Reward Design}}
The base reward is +1 (win), 0 (push), -1 (loss), with +1.5 for blackjack. We add small strategic shaping only where consistent with casino expected value: validated double downs receive multiplicative payouts and invalid doubles are disallowed; successful insurance pays 2:1 but is typically negative expected value and thus rarely selected; early surrender yields -0.5 by design but can be optimal in specific high-risk matchups. This reward shaping encourages robust tactics without biasing toward unrealistic strategies.

\begin{figure*}[tb] 
    \centering
    \includegraphics[width=\linewidth]{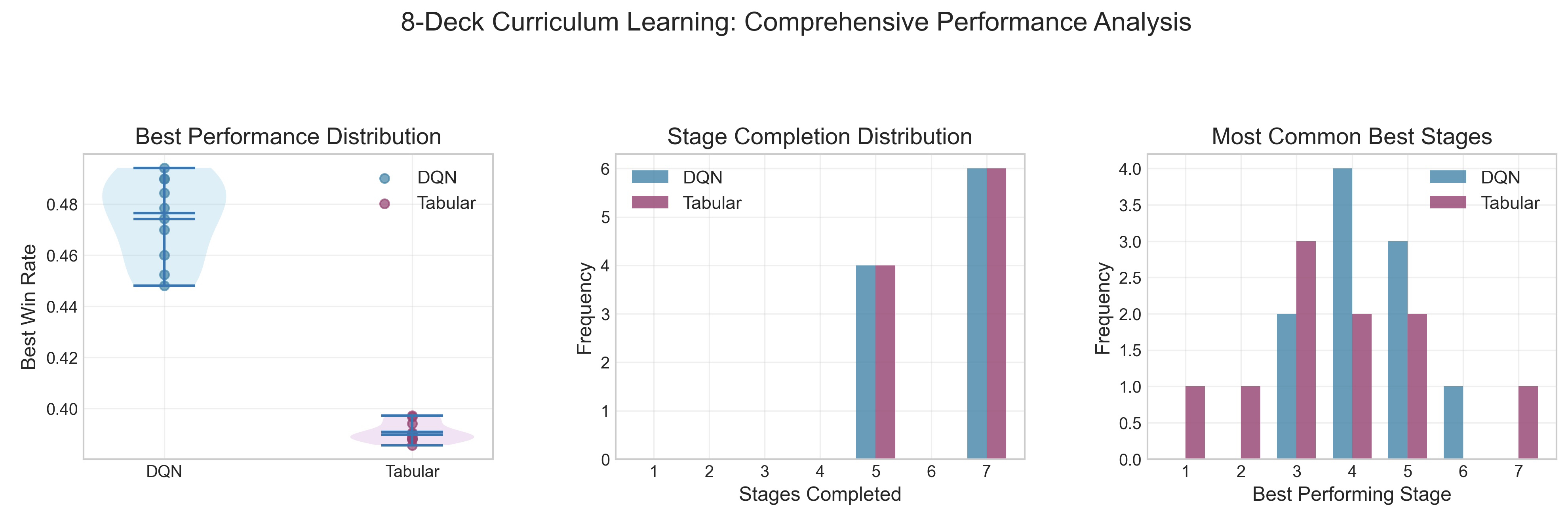}
    \caption{A statistical summary of agent performance over 10 independent runs in the 8-deck curriculum environment.
        The results highlight the DQN agent's significantly higher peak performance distribution (left) and confirm that this peak is most frequently achieved at Stage 4 (right).
        The high stage completion rate for both agents (middle) indicates the curriculum was well-paced and successfully navigated in most runs.}
    \Description{Statistical summary charts of agent performance across 10 runs.}
    \label{fig:curriculum_seeds_part1}
\end{figure*}

\subsection{Agent Setup}
To evaluate the impact of our curriculum across different learning paradigms, we implement and compare two distinct agent architectures. Each agent type is trained independently to solve the Blackjack environment.

\textbf{DQN Agent:} A simple MLP architecture (6→128→128→4) with experience replay buffer size 100,000, target network updates every 1,000 steps, and $\epsilon$-greedy exploration. Learning rate 0.0005 with stage-based adaptation.

\textbf{Tabular Agent:} Direct Q-value lookup with adaptive learning rate of 0.1, frequency-based adjustment, and $\epsilon$-greedy exploration.

\begin{figure*}[t] 
    \centering
    \includegraphics[width=0.8\linewidth]{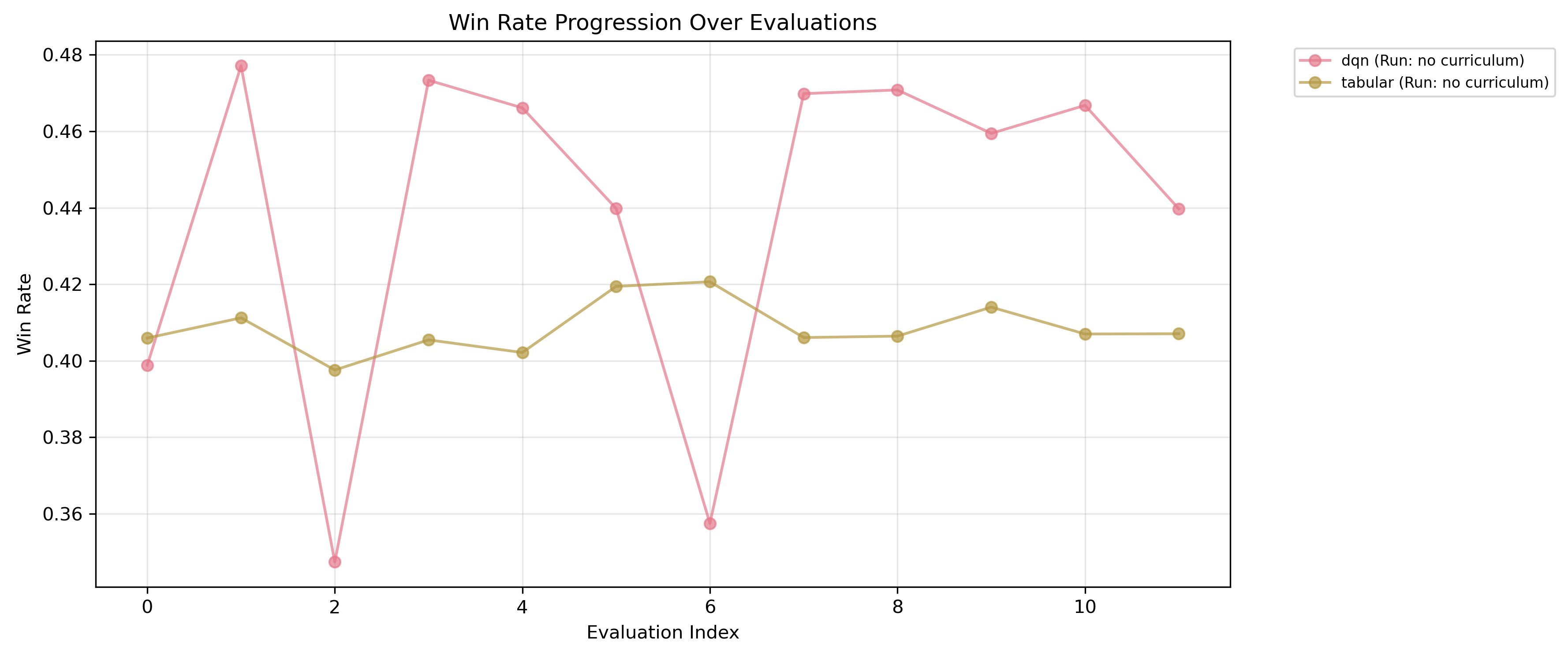}
    \caption{Win rate progression for baseline DQN and Tabular agents trained without a curriculum. The DQN agent's performance is highly volatile, illustrating the instability and exploration challenges that arise when the full action space is introduced at once.}
    \Description{Win rate progression chart for baseline agents showing high volatility.}
    \label{fig:no_curriculum_seeds}
\end{figure*}

\paragraph{\textbf{Adaptive Learning Rate Strategy}}
The system implements adaptive learning rates that adjust based on curriculum stage progression and state visit frequency. For the DQN agent, the learning rate increases by 20\% when advancing to stage 3 or higher, allowing faster adaptation to complex strategies. The tabular agent employs frequency-based learning rate adjustment: $\alpha = \alpha_0 / (1 + \text{visit\_count})$, where $\alpha_0$ is the initial learning rate (0.1) and visit\_count tracks state-action pair frequency. This approach ensures stable learning for frequently visited states while maintaining exploration for rare scenarios.

\paragraph{\textbf{Exploration Strategy and Epsilon Decay}}
Both agents employ $\epsilon$-greedy exploration with exponential decay. The DQN agent uses a decay rate of $0.99995$, while the tabular agent uses $0.9999$ for faster convergence. Epsilon starts at 1.0 and decays to a minimum of 0.05, ensuring a balance between exploration and exploitation throughout training. The decay process allows agents to gradually shift from random exploration to exploiting learned strategies.

\paragraph{\textbf{Baseline}}
Our baseline comparison uses standard training without curriculum learning, where agents have access to all actions from episode 1. Experiments compare this baseline against curriculum learning across multiple deck configurations (infinite, 1-deck, 4-deck, 8-deck with 90\% penetration). Final evaluation uses 100,000 episodes for performance assessment. 

\section{Evaluation}
To assess the effectiveness of our LLM-guided curriculum, we conducted a series of experiments comparing it against a baseline without curriculum learning. Our evaluation is designed to measure improvements in final performance, training efficiency, and learned strategy quality.

\subsection{Experimental Setup}
Our primary analysis focuses on the most realistic casino setting: an 8-deck shoe with 90\% penetration. While we focus on the 8-deck case in the main paper, a comparative analysis for other configurations (1-deck, 4-deck, and infinite-deck) is provided in Appendix \ref{sec:comp_ana}. The baseline for comparison is a standard RL agent trained without a curriculum, meaning it has access to all six actions from the beginning of training.

\begin{figure*}[t] 
    \centering
    \includegraphics[width=0.8\linewidth]{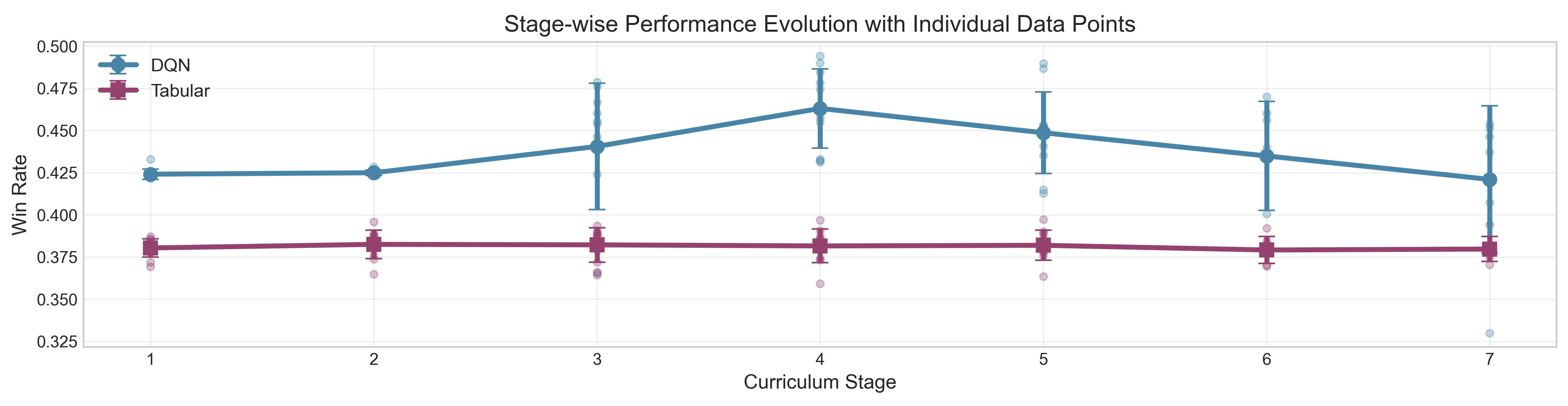}
    \caption{Win rate progression for DQN and Tabular agents trained with the LLM-guided curriculum. The curriculum provides a stable learning trajectory for the DQN agent, preventing the performance volatility seen in baseline training and allowing it to converge to a high and consistent win rate.}
    \Description{Win rate progression chart for curriculum-trained agents showing a stable learning trajectory.}
    \label{fig:curriculum_seeds_part2}
\end{figure*}

\begin{figure*}[t] 
    \centering
    \includegraphics[width=\linewidth]{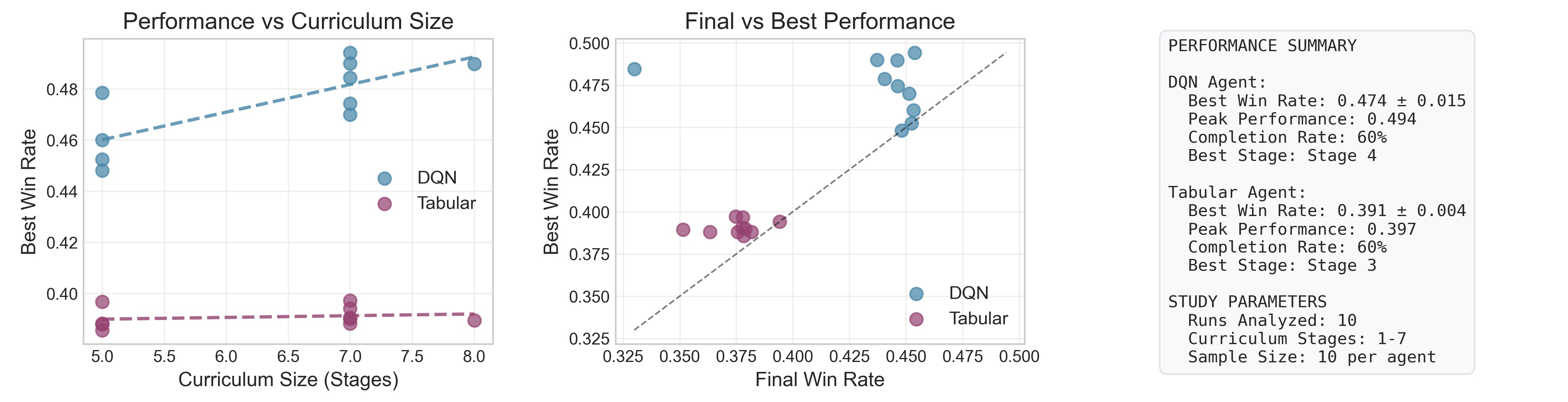}
    \caption{Deeper analysis of agent performance, showing a positive correlation between the number of curriculum stages completed and the best achieved win rate for the DQN agent (left). However, the agent's final performance is often lower than its peak (middle), reinforcing the finding, quantified in the summary (right), that the optimal policy is achieved at an intermediate stage.}
    \Description{Charts showing correlation between completed stages and win rates.}
    \label{fig:curriculum_seeds_part3}
\end{figure*}

\paragraph{\textbf{Training Protocol and Statistical Robustness.}}
For both curriculum and baseline approaches, each agent type (DQN and Tabular) was trained for 500,000 episodes. Final performance was evaluated over a separate 100,000 episodes to ensure stable estimates. To ensure the statistical robustness of our results, all experiments were repeated for 10 independent runs using different random seeds. For each of these 10 curriculum runs, the LLM was engaged dynamically throughout the training process. After the completion of each stage, a performance summary was sent to the LLM, which then determined whether the agent should advance. This per-run, adaptive interaction tests the robustness of our LLM-as-a-coach framework.

\paragraph{\textbf{Performance Metrics.}}
We evaluate agent performance using several metrics: final win rate, average per-episode reward, and bust rate. We also analyze training efficiency by measuring wall-clock training time and the number of episodes required to reach peak performance. Finally, we conduct a qualitative analysis of the agents' learned policies using strategy heatmaps.

\paragraph{\textbf{Reproducibility.}}
To ensure our work is fully reproducible, we release all source code, experiment scripts, and configuration files. We also provide the JSON logs for every training and evaluation run, which include explicit random seeds, episode budgets, and stage thresholds. All figures and tables in this paper are generated directly from these logs using our public analysis scripts, available at: 
\url{https://github.com/LUH-AI-devnerds/llm-guided-curriculum-rl}.

\subsection{Quantitative Results}

Our experimental results show that the LLM-guided curriculum provides significant advantages in both final performance and training efficiency, particularly for the DQN agent.

As shown in Figure~\ref{fig:curriculum_seeds_part1}, the curriculum-trained DQN agent consistently achieves a higher win rate distribution. On average, its best win rate was 47.41\%, a substantial improvement over the 43.97\% achieved by the baseline agent (shown in Figure~\ref{fig:no_curriculum_seeds}). This performance gain was accompanied by a significant reduction in risk-taking, as the average bust rate fell from 32.9\% for the baseline to 28.0\% for the curriculum agent.

\begin{table}[htbp]
    \centering
    \caption{Win rate by curriculum stage (8-deck). The values represent the best observed win rate for a single run at each stage.}
    \label{tab:stage_results}
    \begin{tabular}{lrrrr}
        \toprule
        Stage & Available Actions & DQN (\%) & Tabular (\%) \\
        \midrule
        1: Hit/Stand & 0,1 & 42.24 & 39.74 \\
        2: +Double & 0,1,2 & 42.37 & 38.72 \\
        3: +Split (no Double) & 0,1,3 & 45.54 & 38.78 \\
        4: Full Basic (0--3) & 0,1,2,3 & \textbf{49.42} & 38.15 \\
        5: +Insurance & 0,1,2,3,5 & 48.66 & 40.00 \\
        6: +Surrender & 0,1,2,3,4 & 46.57 & 37.78 \\
        7: All Actions & 0,1,2,3,4,5 & 44.63 & 37.90 \\
        \bottomrule
    \end{tabular}
\end{table}

\begin{figure}[htbp]
    \centering
    \includegraphics[width=\linewidth]{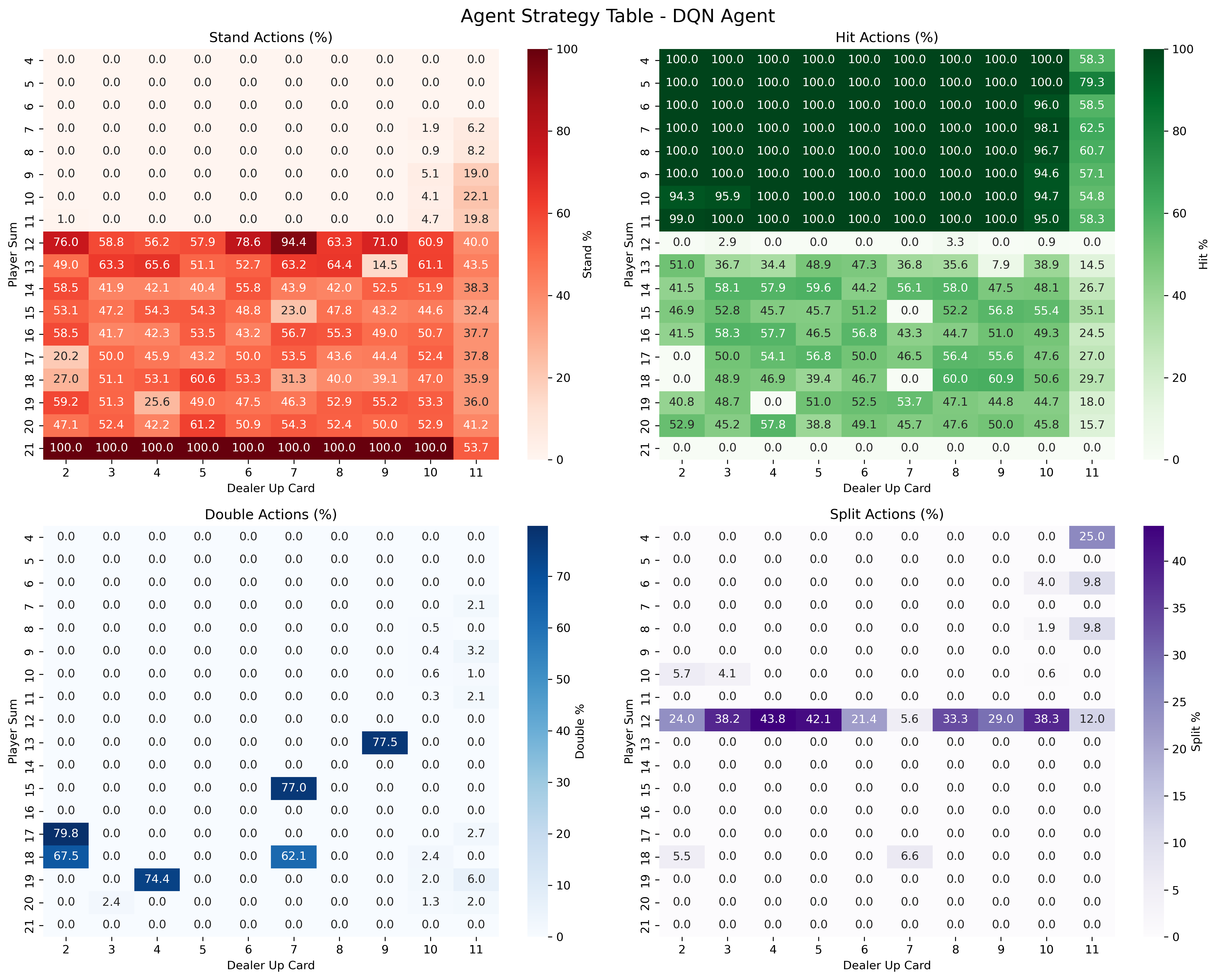}
    \caption{A strategy heatmap for an agent trained without a curriculum. Rows are the player's sum, columns are the dealer's up card. Cell colors represent the chosen action. The strategy appears less structured than the curriculum-trained agent.}
    \Description{Strategy heatmap showing unstructured decision boundaries for the baseline agent.}
    \label{fig:heatmap_nocurr}
\end{figure}

\paragraph{\textbf{Training Efficiency.}}
The curriculum dramatically improves agent efficiency for both architectures. The DQN agent, in particular, saw a significant acceleration: the curriculum-trained agents completed their entire training protocol in an average of only \textbf{12.52 minutes}. In contrast, the baseline agent's average training time was approximately 48.4 minutes. The fact that the curriculum agent can be fully trained in a fraction of the time it takes to merely evaluate the baseline demonstrates a substantial improvement in the overall workflow. Efficiency gains were also clear for the simpler Tabular Q-learning agent, which saw its average training time reduced from one minute for the baseline approach to just \textbf{40 seconds} with the curriculum.

\begin{figure}[htbp]
    \centering
    \includegraphics[width=\linewidth]{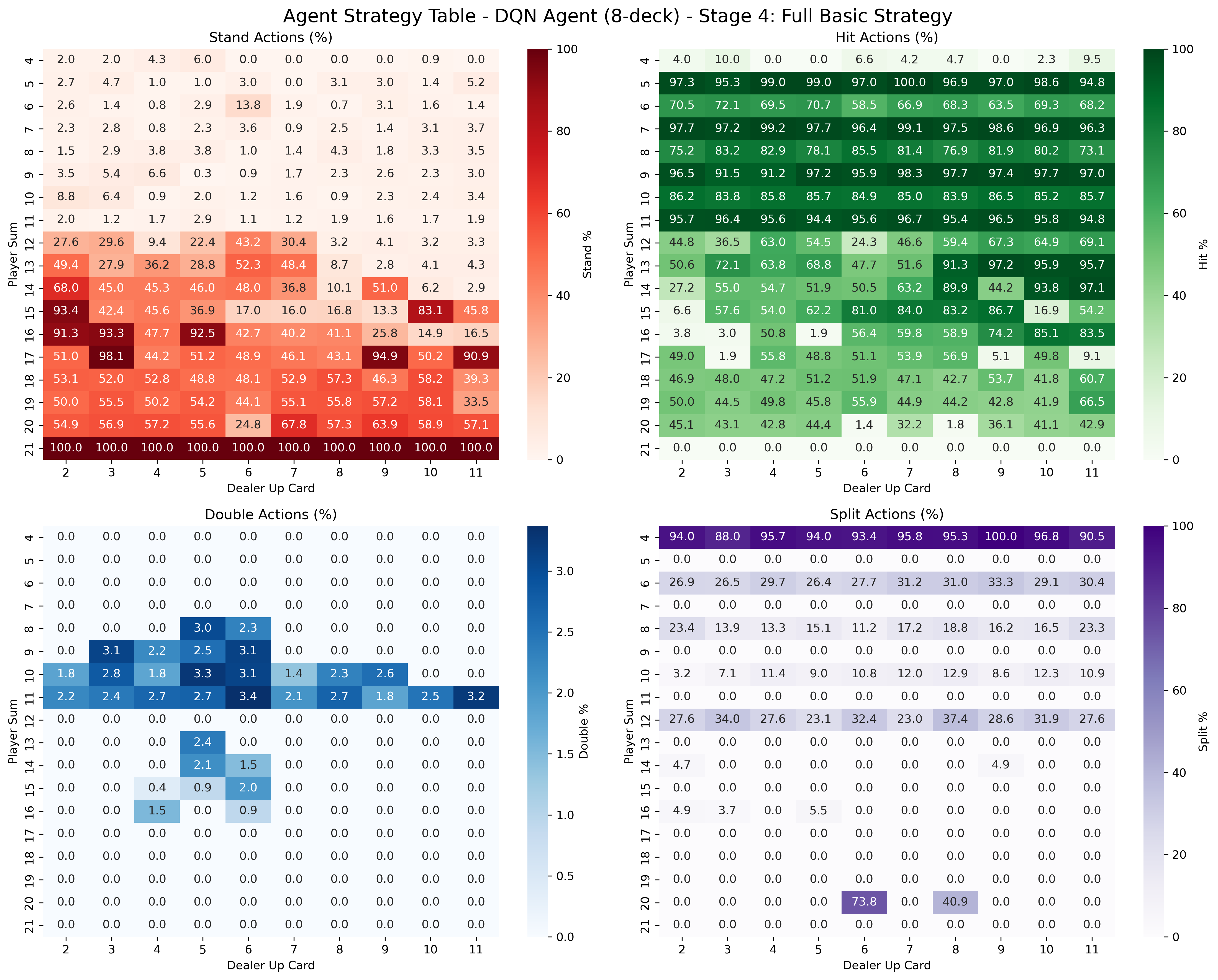}
    \caption{A strategy heatmap for an agent trained with a curriculum. Rows are the player's sum, columns are the dealer's up card. Cell colors represent the chosen action. The strategy appears more structured than the no curriculum-trained agent.}
    \Description{Strategy heatmap showing structured decision boundaries for the curriculum agent.}
    \label{fig:strategy_table_heatmap_stage4}
\end{figure}

\paragraph{\textbf{Stage-wise Analysis.}}

The stage-wise evolution plot in Figure~\ref{fig:curriculum_seeds_part2} and the data in Table~\ref{tab:stage_results} clearly show the agent's performance progressively improving and then peaking. Agents most frequently achieve their peak performance at Stage 4 ("Full Basic"), a finding supported by the distribution shown in the top-right panel of Figure~\ref{fig:curriculum_seeds_part1}. With this full basic strategy available, our DQN agent reaches a win rate of an impressive $49.42\%$. Performance often stabilizes or slightly degrades in later stages. This is further confirmed by our comparison of final and all-time best win rates in Figure~\ref{fig:curriculum_seeds_part3}, where most points lie below the diagonal, indicating the final performance was often lower than the peak. We theorize that the reason for performance decay later on is the high complexity of the Insurance and Surrender actions. 
These are only required in specialized strategies with otherwise high risks, making them much harder to apply.
Even in later stages of the curriculum, however, the win rate stays above the baseline win rate, only falling as low as $44.63\%$.

\paragraph{\textbf{DQN and Tabular Comparison.}}

Figure~\ref{fig:curriculum_seeds_part3} illustrates the performance gap between the two agent types. As detailed in Table~\ref{tab:stage_results}, the Tabular agent's peak win rate was only 49\%, significantly lower than the DQN's peak of 49.42\%. The Tabular agent's limited capacity prevents it from learning the nuanced, state-conditional strategies that the DQN's function approximator can capture, resulting in a less effective final policy with a higher bust rate. 
Even in this case, however, the curriculum is useful.
This demonstrates that the complexity of the learning algorithm is less of a factor in the success of our curriculum than the complexity of the environment.

\subsection{Qualitative Strategy Analysis}

To understand how the curriculum improves performance, we visualized the learned policies of the DQN agents using strategy heatmaps. This qualitative analysis reveals a strong difference in the learned strategies.

The agent trained without a curriculum learns a hesitant and unstructured policy (Figure \ref{fig:heatmap_nocurr}). Its strategy map is diffuse, concentrating almost exclusively on hitting and standing. 
We can see that the main deciding factor between hitting and standing is the player's sum, where at a value of 12, the agent switches to predominantly standing compared to hitting below 12. The dealer's sum does not play a large role in this decision. It also fails to learn the proper conditions for using high-value tactical actions like Double Down and Split, even in obviously advantageous situations (e.g., a player's hand of 11 compared to the dealer's 6). This leads to under-utilization of good opportunities and over-hitting in marginal situations.

In contrast, the agent trained with the curriculum learns a crisp, coherent policy that closely resembles established Blackjack basic strategy (Figure \ref{fig:strategy_table_heatmap_stage4}). The staged introduction of actions allows it to master each one in turn. Its final strategy map shows clear, well-defined regions for standing (e.g., on 12-16 for a dealer's bust card), doubling down (e.g., on hard 10-11), and splitting pairs (e.g., Aces and 8s). This confirms that the curriculum does not merely boost metrics but fundamentally guides the agent toward learning a more sophisticated and strategically sound decision-making process.

\subsection{Limitations and Future Work}
Our study, while demonstrating the promise of LLM-guided curricula for Blackjack, has several limitations that suggest avenues for future work.

The proposed action curriculum approach is highly effective for environments like Blackjack with a discrete, well-defined, and separable action space. However, the applicability of this specific method to other domains is an open question. Its effectiveness may be limited in environments with continuous action spaces (e.g., robotics) or where actions are not easily isolated into meaningful subsets of increasing complexity. Future research should explore adapting this concept to a broader range of RL problems.

\paragraph{\textbf{Dependence on a Specific LLM.}}
Our framework relies on the Gemini-2.0-flash-001 model accessed via a public API. This introduces two potential threats to validity. First, since the underlying model can be updated by its provider at any time, the strict reproducibility of the generated curriculum cannot be guaranteed indefinitely. Second, our findings are specific to this particular LLM. Different models (e.g., GPT-4, Llama) may possess different reasoning capabilities and inherent biases, which could lead them to generate substantially different and potentially more or less effective curricula.

\paragraph{\textbf{Curriculum Monotonicity.}}
Our results indicate that agent performance peaks at an intermediate stage of the curriculum (Stage 4), after which the introduction of more situational actions leads to a decline in win rate. A limitation of our current adaptive framework is its monotonic nature: it only allows the agent to progress forward with the goal of eventually using all actions. 
It does not include a mechanism to determine if settling on a restricted action space indefinitely is a better choice. 
An extended version of our curriculum could contain such an option.

\paragraph{\textbf{Scope of the Simulation.}}
Finally, while our 8-deck environment is a realistic simulation of the card game, it does not capture all aspects of real-world casino play. We do not model betting strategies or dynamic table limits, which are crucial for managing bankroll and can influence the expected value of actions like Insurance and Early Surrender, especially when combined with card counting.
Since our results are very promising for basic strategy play, a deeper setting supporting such advanced play could prove to be an interesting future avenue for research.

\section{Conclusion}
This work presents a novel LLM-guided action curriculum learning framework for reinforcement learning in the Blackjack domain, demonstrating the effectiveness of combining language model intelligence with structured curriculum design. 
We find that introducing actions one by one based on an LLM plan allows Q-Learning agents to reach expert-level win rates in realistic 8-deck settings, surpassing non-curriculum variants both in terms of performance and sample efficiency.

These results show that the system effectively manages progressive complexity introduction through adaptive success thresholds and learns meaningful policies corresponding to real Blackjack play.

While these are impressive results, our setting does not yet integrate explicit card-counting and betting decisions to study bankroll growth under risk constraints. Furthermore, we use casino rules, but Blackjack also features richer rule sets like "dealer hits soft 17". These could be valuable extensions for studying RL on Blackjack.
We also believe that our action-based curriculum extends to a broader set of tactical decision-making tasks beyond Blackjack. Especially with further extensions like adaptive thresholds, action-based curricula could enable efficient task decomposition and learning across many complex RL tasks.

\bibliographystyle{ACM-Reference-Format}
\bibliography{Bib/Lib}

\appendix
\onecolumn 

\section{Visualizations of Agent Performance and Strategy}

\subsection{Agent Progression Through Curriculum Stages}
\begin{figure}[H]
    \centering
    \includegraphics[width=0.9\linewidth, keepaspectratio]{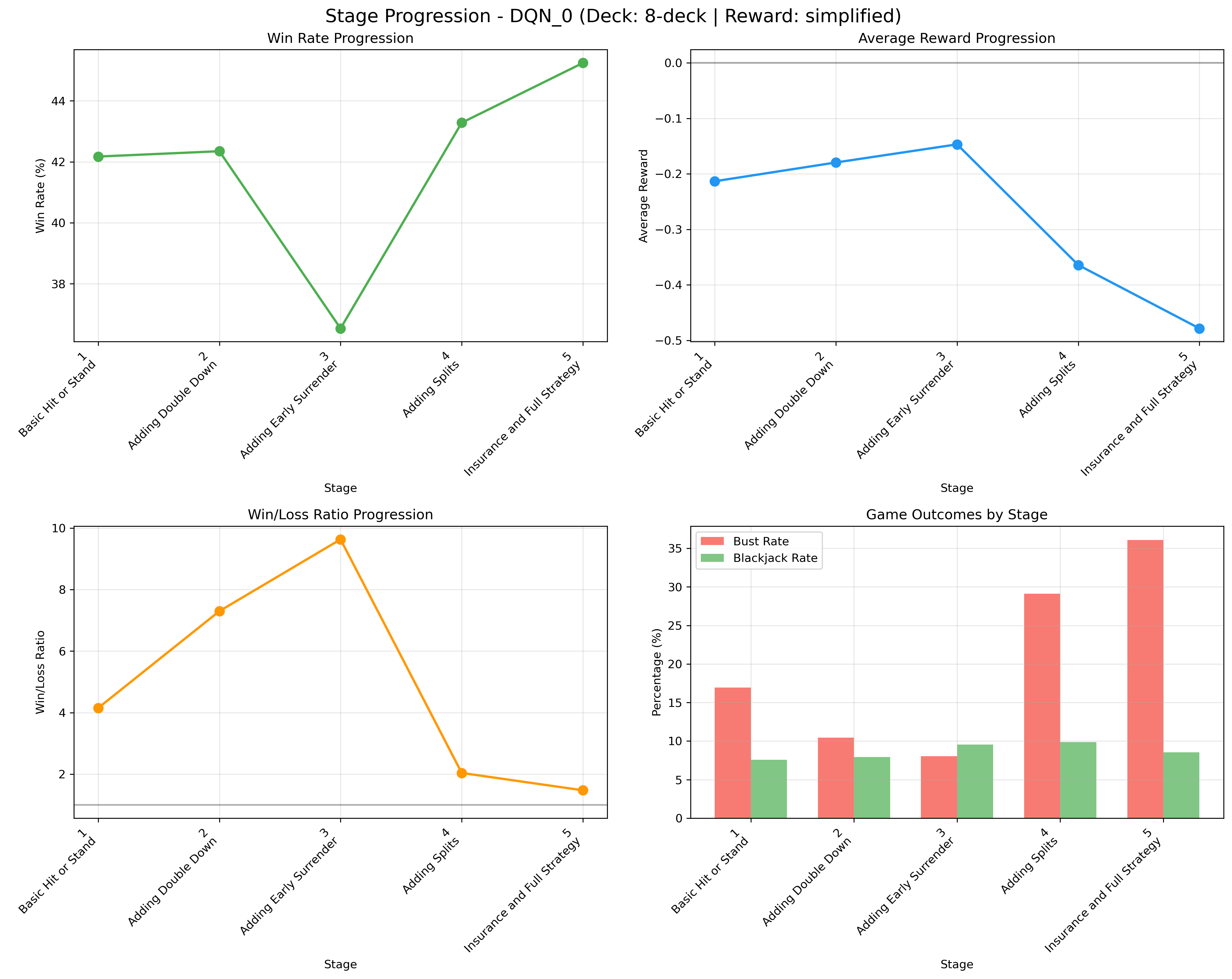}
   \caption{Performance progression of the DQN agent through a 5-stage curriculum in an 8-deck game.}
   \Description{DQN agent performance progression through 5 curriculum stages.}
   \label{fig:stage_progression}
\end{figure}

Figure \ref{fig:stage_progression} illustrates the performance progression of the DQN agent. A notable dip in win rate is observable at Stage 3, characteristic of learning the complex "Early Surrender" action.

\subsection{Comparative Analysis Across Deck Configurations}
\label{sec:comp_ana}
The following figures compare the DQN agent against the Tabular baseline across different deck sizes. Note the superior win rates and lower bust rates for the DQN agent.

\begin{figure}[H]
    \centering
    \includegraphics[width=1.0\linewidth, height=0.42\textheight, keepaspectratio]{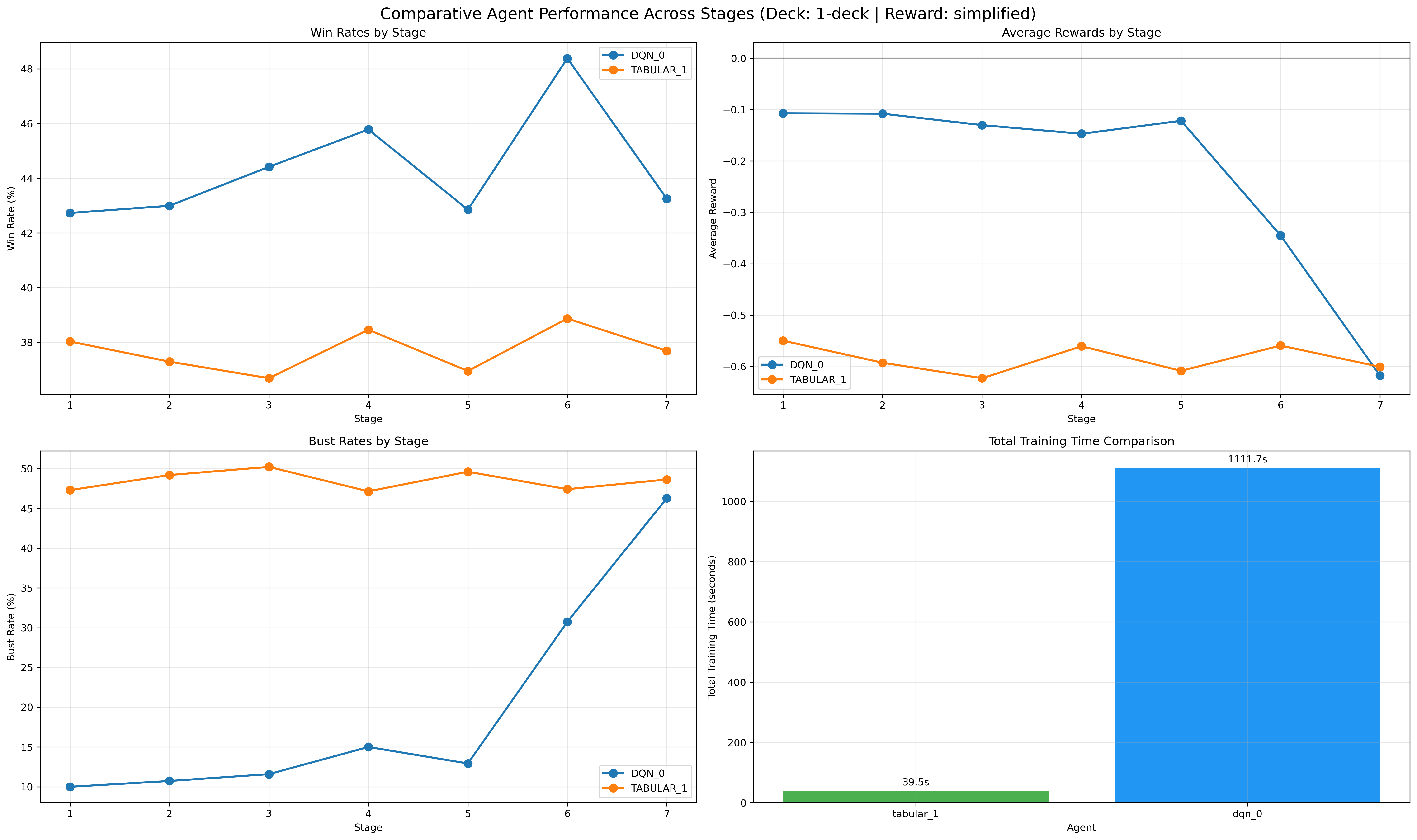}
   \caption{Comparative analysis in a 1-deck environment.}
   \Description{Comparative analysis chart for 1-deck environment.}
   \label{fig:1-deck}
\end{figure}

\begin{figure}[H]
    \centering
    \includegraphics[width=1.0\linewidth, height=0.42\textheight, keepaspectratio]{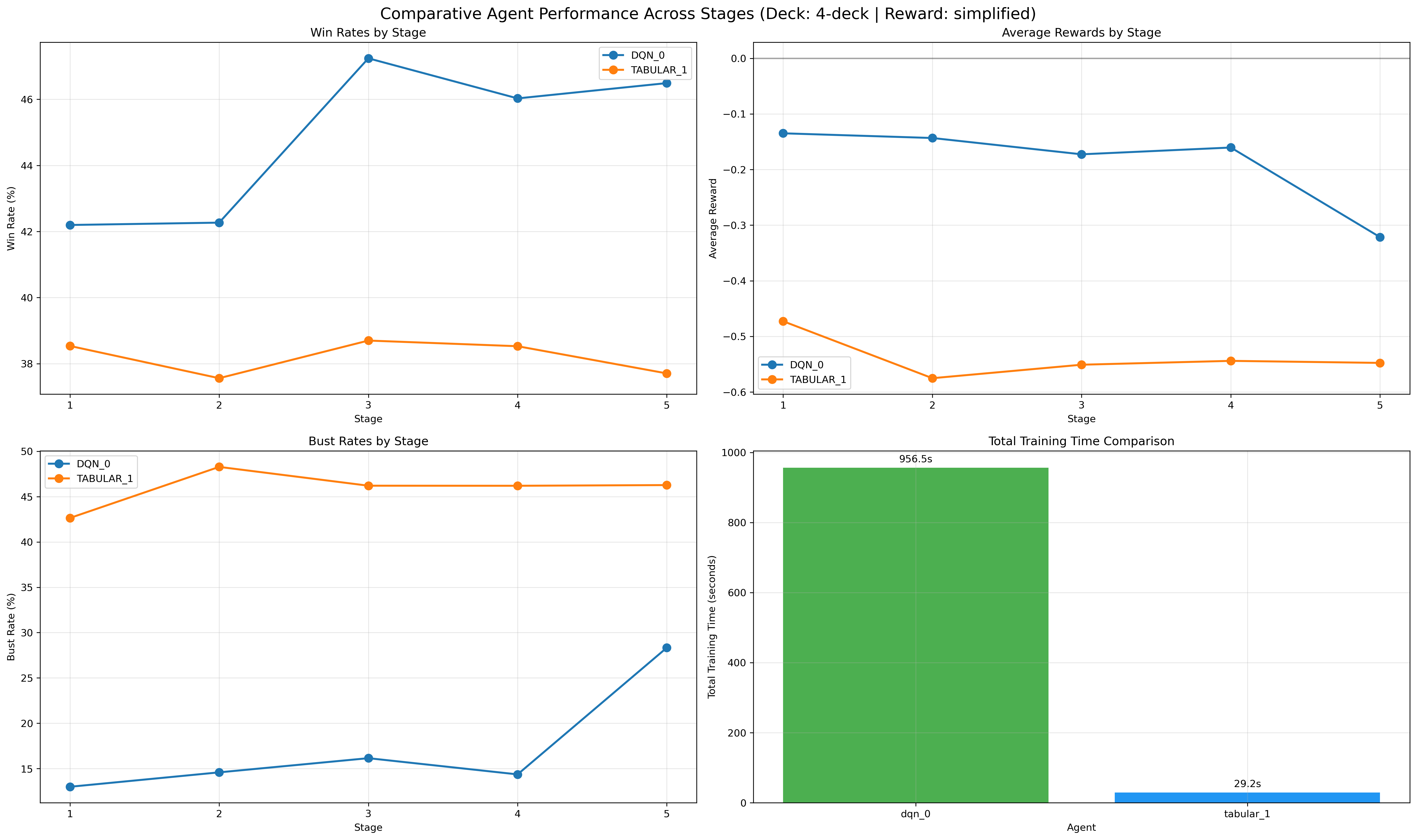}
   \caption{Comparative analysis in a 4-deck environment.}
   \Description{Comparative analysis chart for 4-deck environment.}
   \label{fig:4-deck}
\end{figure}

\begin{figure}[H]
    \centering
    \includegraphics[width=1.0\linewidth, height=0.42\textheight, keepaspectratio]{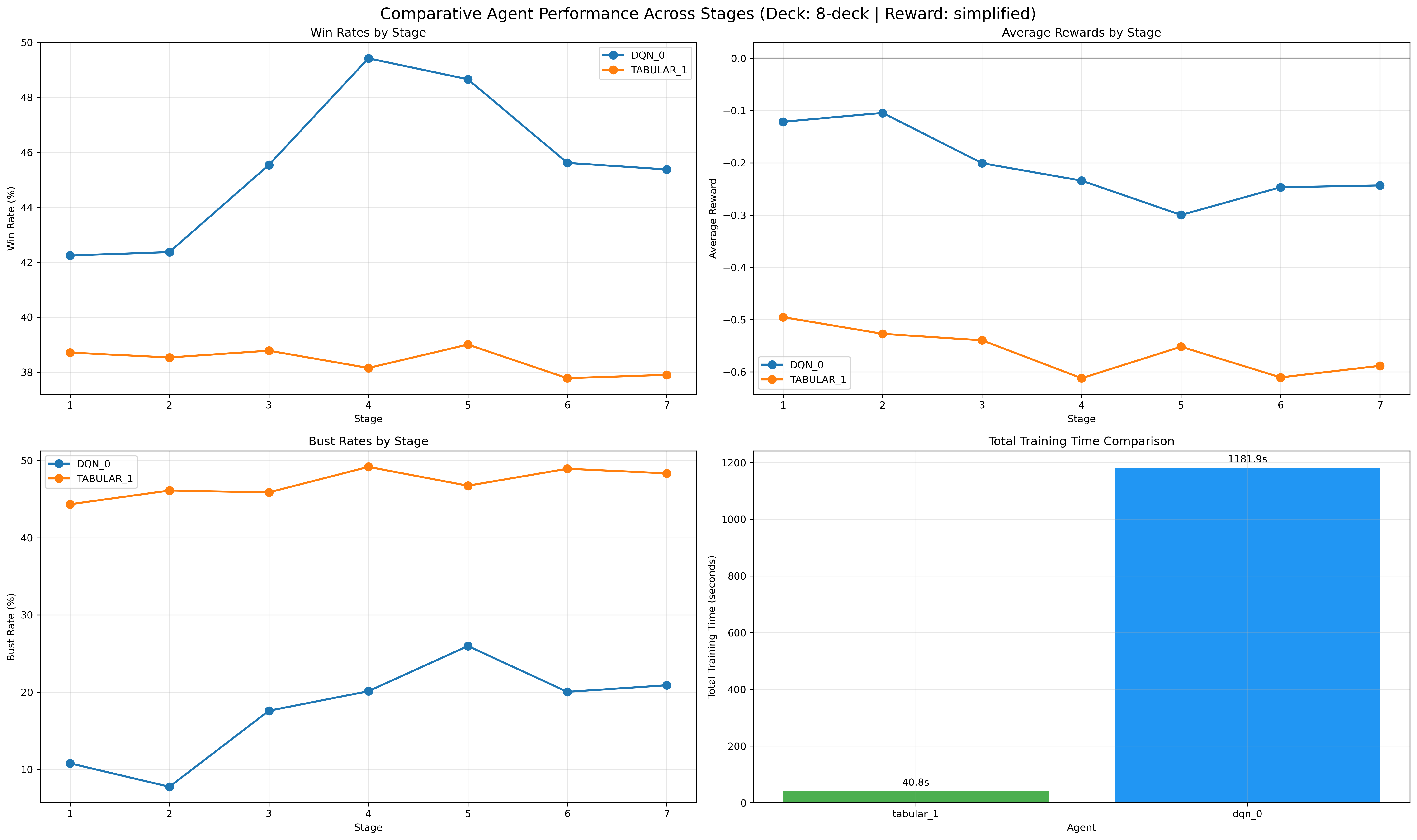}
   \caption{Comparative analysis in an 8-deck environment.}
   \Description{Comparative analysis chart for 8-deck environment.}
   \label{fig:8-deck}
\end{figure}

\begin{figure}[H]
    \centering
    \includegraphics[width=1.0\linewidth, height=0.42\textheight, keepaspectratio]{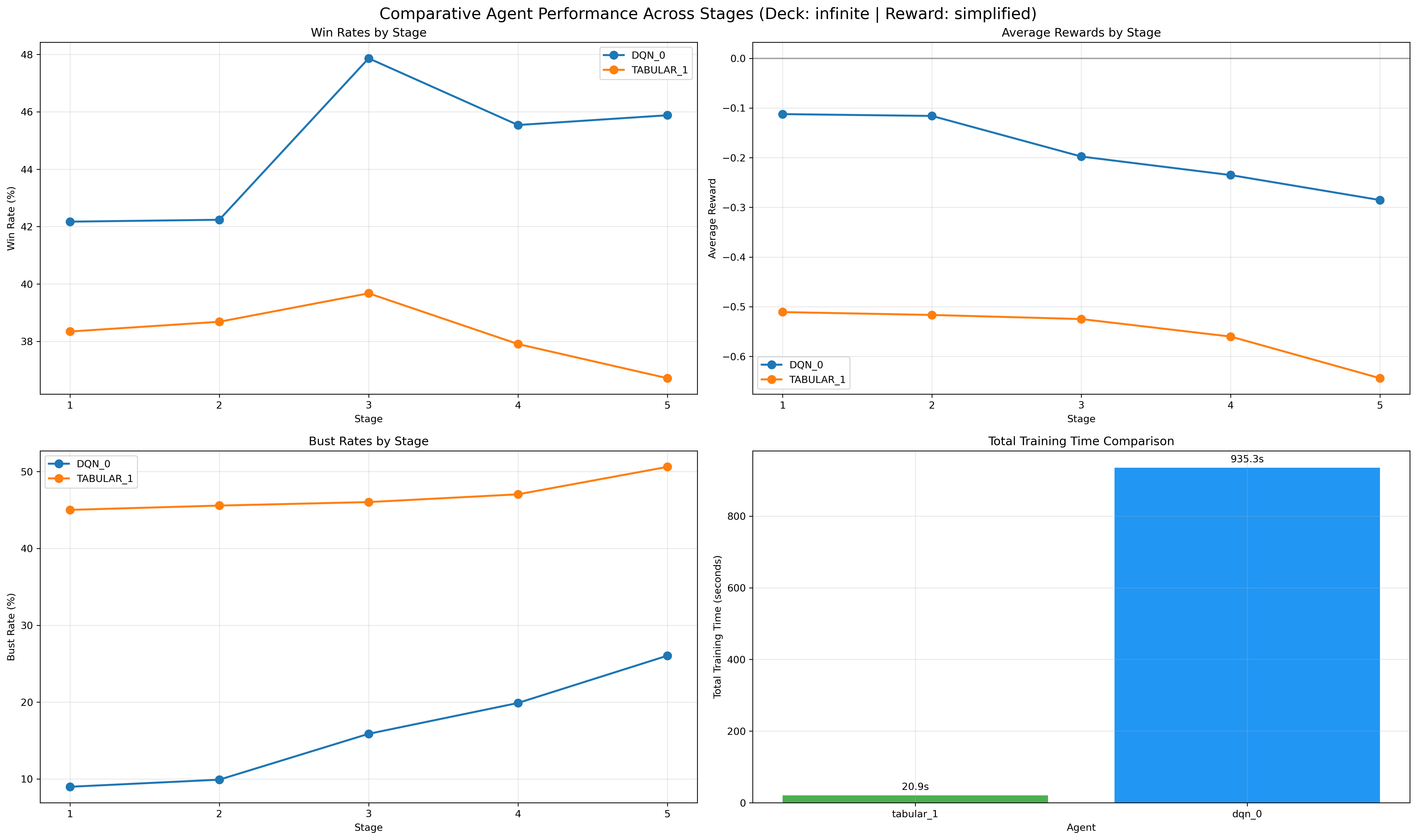}
   \caption{Comparative analysis in an infinite-deck environment.}
   \Description{Comparative analysis chart for infinite-deck environment.}
   \label{fig:inf-deck}
\end{figure}

\subsection{Learned Policy Visualization}
To validate the strategy, we visualize the learned policy at Stage 3.

\begin{figure}[H]
    \centering
    \includegraphics[width=0.9\linewidth, height=0.45\textheight, keepaspectratio]{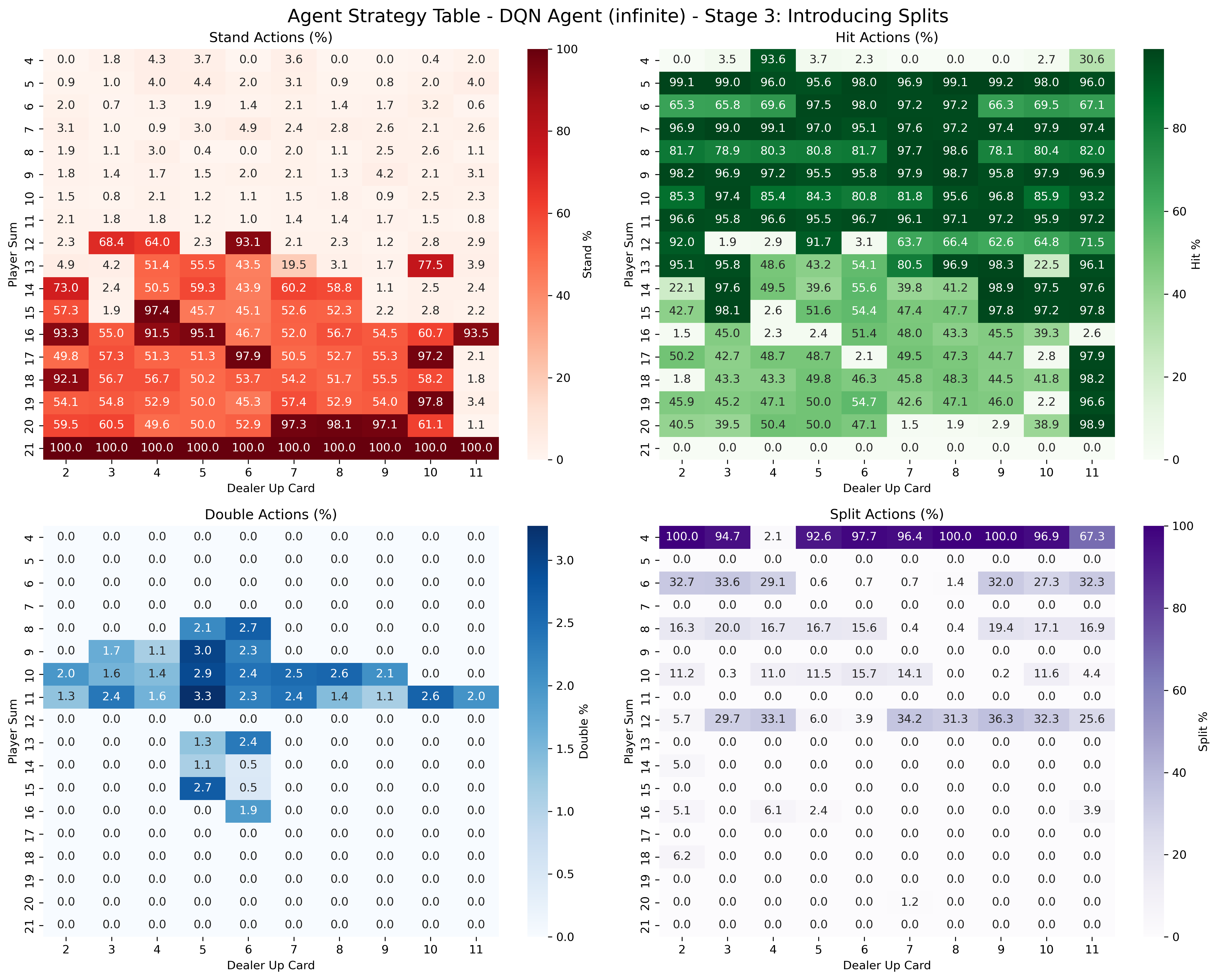}
   \caption{Heatmap visualization of the DQN agent's learned policy at Stage 3 ("Introducing Splits").}
   \Description{Heatmap of learned policy at Stage 3.}
   \label{fig:strategy_table_heatmap_appendix}
\end{figure}

\section{Prompting for Curriculum Generation}
\subsection{Curriculum Generation Prompt}
\noindent
\begin{lstlisting}[breaklines=true, basicstyle=\small\ttfamily]
You are an expert in reinforcement learning curriculum design for Blackjack. 
    
    Available actions:
       0: "Stand - Stay with current hand",
        1: "Hit - Draw another card",
        2: "Double Down - Double bet and draw one card",
        3: "Split - Split pair into two hands",
        4: "Early Surrender - Give up hand and lose half bet",
        5: "Insurance - Bet dealer has blackjack when showing Ace",
    
    Design the number of stages that is needed to train RL agents in Blackjack.
    Each stage should gradually increase complexity and introduce new actions.
    
    For each stage, specify:
    1. Stage name
    2. Available actions (list of action indices: 0, 1, 2, 3, 4, 5)
    3. Description of learning objectives
    4. Difficulty level (1-3 scale)
    5. Success threshold (win rate 0.0-1.0 to advance)
    
    You arent forced to use actions in the order of the list. You can use actions in any order you want.
    You can even choose to not use an action at all.
    The complexity of actions are as follows:
    - Stand: 1
    - Hit: 1
    - Early Surrender: 2
    - Insurance: 2
    - Double Down: 3
    - Split: 3
    Respond in this exact JSON format:
    {{
        "stages": [
            {{
                "stage_id": 1,
                "name": "Stage Name",
                "available_actions": [0, 1],
                "description": "Learning objectives...",
                "difficulty": 1,
                "success_threshold": 0.35
            }},
            ...
        ],
        "rationale": "Explanation of curriculum design..."
    }}
    
    Focus on progressive skill building and realistic success thresholds.
    Consider that early surrender (action 4) and insurance (action 5) are advanced strategies
    that should be introduced in later stages after basic strategy is mastered.
\end{lstlisting}

\subsection{Curriculum Adaptation Prompt}
\begin{lstlisting}[breaklines=true, basicstyle=\small\ttfamily]
You are analyzing an RL agent's performance in Blackjack curriculum learning.
        
        Current Stage: {current_stage.name}
        Available Actions: {current_stage.available_actions}
        Stage Description: {current_stage.description}
        Success Threshold: {current_stage.success_threshold}
        
        Agent Performance:
        - Win Rate: {performance_summary['win_rate']:.3f}
        - Average Reward: {performance_summary['average_reward']:.3f}
        - Episodes Trained: {performance_summary['episodes_trained']}
        - Struggling with actions: {performance_summary['struggling_actions']}
        
        Action Descriptions:
        {self.action_descriptions}
        
        Based on this performance, recommend:
        1. Should the agent advance to next stage? (yes/no)
        2. What actions should be emphasized in next training episodes?
        3. Any curriculum modifications needed?

        You arent forced to use actions in the order of the list. You can use actions in any order you want.
        You can even choose to not use an action at all.
        The complexity of actions are as follows:
        - Stand: 1
        - Hit: 1
        - Early Surrender: 3
        - Insurance: 3
        - Double Down: 2
        - Split: 2
        
        Respond in JSON format:
        {{
            "advance_stage": true/false,
            "recommended_actions": [list of action indices to focus on],
            "curriculum_modifications": {{
                "adjust_threshold": 0.XX,
                "add_actions": [list],
                "remove_actions": [list]
            }},
            "reasoning": "Explanation of recommendations..."
        }}
\end{lstlisting}

\subsection{Example Generated Curriculum}
\begin{lstlisting}[language=Python, breaklines=true, basicstyle=\small\ttfamily]
{
    "curriculum_stages": [
        {
            "stage_id": 1,
            "name": "Basic Hit & Stand",
            "available_actions": [
                0,
                1
            ],
            "description": "Learn the fundamental actions of hitting and standing. Agent should learn to hit when the hand total is low and stand when it's high, based on the dealer's upcard. Focus on memorizing the most basic actions for each hand.",
            "difficulty": 1,
            "success_threshold": 0.4
        },
        {
            "stage_id": 2,
            "name": "Introduction to Doubling Down",
            "available_actions": [
                0,
                1,
                2
            ],
            "description": "Introduce the 'Double Down' action. Agent should learn to recognize advantageous situations for doubling down (e.g., hard 11, hard 10 against low dealer upcards).",
            "difficulty": 2,
            "success_threshold": 0.42
        },
        {
            "stage_id": 3,
            "name": "Splitting Pairs",
            "available_actions": [
                0,
                1,
                3
            ],
            "description": "Introduce the 'Split' action. Focus on learning when to split pairs (e.g., splitting aces and eights almost always, but learn to avoid splitting 10s). The game is simplified to avoid 'Double Down' to allow better focus on the newly introduced action.",
            "difficulty": 3,
            "success_threshold": 0.43
        },
        {
            "stage_id": 4,
            "name": "Full Basic Strategy",
            "available_actions": [
                0,
                1,
                2,
                3
            ],
            "description": "Combine all basic actions (Hit, Stand, Double Down, and Split). Agent must now master a comprehensive basic strategy, choosing the optimal action in various situations. This stage will require the agent to balance all the actions they know, and use them efficiently. It is not expected to become an expert, but to understand how each action should be used in different circumstances.",
            "difficulty": 4,
            "success_threshold": 0.45
        },
        {
            "stage_id": 5,
            "name": "Insurance Decision",
            "available_actions": [
                0,
                1,
                2,
                3,
                5
            ],
            "description": "Introduce 'Insurance'. The agent should learn that insurance is generally a bad bet except in very specific card counting scenarios, which are not available now. Agent should learn to almost always decline Insurance. This stage is about learning the expected value of this advanced action.",
            "difficulty": 4,
            "success_threshold": 0.45
        },
        {
            "stage_id": 6,
            "name": "Early Surrender",
            "available_actions": [
                0,
                1,
                2,
                3,
                4
            ],
            "description": "Introduce the 'Early Surrender' action. The agent should learn when early surrender is advantageous, which typically involves giving up unfavorable hands against a strong dealer upcard.",
            "difficulty": 4,
            "success_threshold": 0.46
        },
        {
            "stage_id": 7,
            "name": "All Actions Master",
            "available_actions": [
                0,
                1,
                2,
                3,
                4,
                5
            ],
            "description": "The agent can now choose any action in all situations. Mastering all actions requires the agent to intelligently combine all strategies learned to create a holistic and performant strategy.",
            "difficulty": 5,
            "success_threshold": 0.47
        }
    ]
}
\end{lstlisting}

\end{document}